\documentclass[runningheads]{llncs}

\usepackage{eccv}

\usepackage{eccvabbrv}

\usepackage{graphicx}
\usepackage{booktabs}
\usepackage{multirow}
\usepackage{algorithm}
\usepackage{algorithmic}
\usepackage[table]{xcolor}
\usepackage{wrapfig}
\usepackage{booktabs}
\usepackage{subcaption}
\usepackage[accsupp]{axessibility}  

\usepackage{hyperref}

\usepackage{orcidlink}

\begin{document}

\title{ScalePredictor: Instance-aware Scale Learning for Accurate Quantization of Vision Transformers} 

\titlerunning{Abbreviated paper title}

\author{Changjun Li\inst{1}\thanks{Equal contribution.} \and
Runqing Jiang\inst{1}$^*$ \and
Lian Xu\inst{2} \and
Ye Zhang\inst{1} \and
Qingyong Hu\inst{3} \and
Yulan Guo\inst{1}}

\authorrunning{C.~Li et al.}

\institute{School of Electronics and Communication Engineering, Shenzhen Campus, Sun Yat-sen University, Shenzhen, China \\
\email{\{lichj68, jiangrq3\}@mail2.sysu.edu.cn, \{zhangy2658, guoyulan\}@sysu.edu.cn} \and
Computer Science and Software Engineering, The University of Western Australia, Perth, Australia \and
University of Oxford, Oxford, United Kingdom}
\maketitle

\begin{abstract}
Vision Transformers have achieved remarkable success in many fields, yet their deployment on edge devices remains challenging due to their substantial computational demands. Post-Training Quantization (PTQ) offers an attractive solution by compressing models using a small calibration set with minimal training overhead. However, most existing PTQ works adopt a static quantization paradigm that is uniformly applied to all instances. Given the substantial diversity of natural images, the activation distributions vary significantly across samples, making these methods inherently suboptimal. In this paper, we propose ScalePredictor, a dynamic quantization framework for accurate and efficient quantization scale learning of ViTs. We first reveal a hidden correlation between the distribution range of shallow-layer activations and the optimal scales of deeper layers. Based on this, we develop a scale learning mechanism that integrates an efficient range extraction approach to capture robust range statistics at the shallow stage, which are then fed into a Taylor-motivated polynomial scale projection module to generate all quantization scales simultaneously. With the efficiency of polynomial approximation, ScalePredictor introduces insignificant computational overhead while avoiding costly just-in-time calibration. Extensive experiments on ImageNet demonstrate that ScalePredictor consistently outperforms prior PTQ methods, achieving a more favorable accuracy–efficiency trade-off. Code and additional results are shown in the supplementary materials.
  
  \keywords{Vision Transformers \and Quantization \and Dynamic Network}
\end{abstract}

\section{Introduction}
\label{sec:intro}
The rapid advancement of Vision Transformers (ViTs) has substantially pushed the performance boundaries of a wide range of computer vision tasks~\cite{chen2021crossvit,liu2021swin,yuan2021t2tvit,li2022exploring,fang2021you,xu2024mctformerplus}. Despite their impressive success, the complex architectures and intensive computations of ViTs pose significant challenges for deployment on resource-constrained edge devices. To alleviate these limitations, a variety of model compression techniques have been developed, including compact architectural design~\cite{chen2022mobile,li2022efficientformer}, network pruning~\cite{jiang2022sparse,molchanov2019importance}, knowledge distillation~\cite{jiang2023knowledge,yang2024dual}, and model quantization~\cite{jiang2025aiqvit,li2025pack}. Among these approaches, model quantization has attracted considerable attention, which converts model weights and intermediate activations from high-precision floating-point representations (\eg, 32-bit) to low-bit formats (\eg, 8-bit).  As a lightweight quantization paradigm, post-training quantization (PTQ) methods compress neural networks without the need for the entire training data and expensive retraining procedures, making it attractive for real-world deployment, especially for modern ViTs~\cite{jiang2025aiqvit,li2025pack,zhong2023s,wu2025fima}. 
\begin{figure}[!t]
  \centering
  \begin{minipage}{0.317\textwidth}\centering\scriptsize(a)\end{minipage}%
  \begin{minipage}{0.317\textwidth}\centering\scriptsize(b)\end{minipage}%
  \begin{minipage}{0.317\textwidth}\centering\scriptsize(c)\end{minipage}\\[2pt]
  \includegraphics[width=\textwidth]{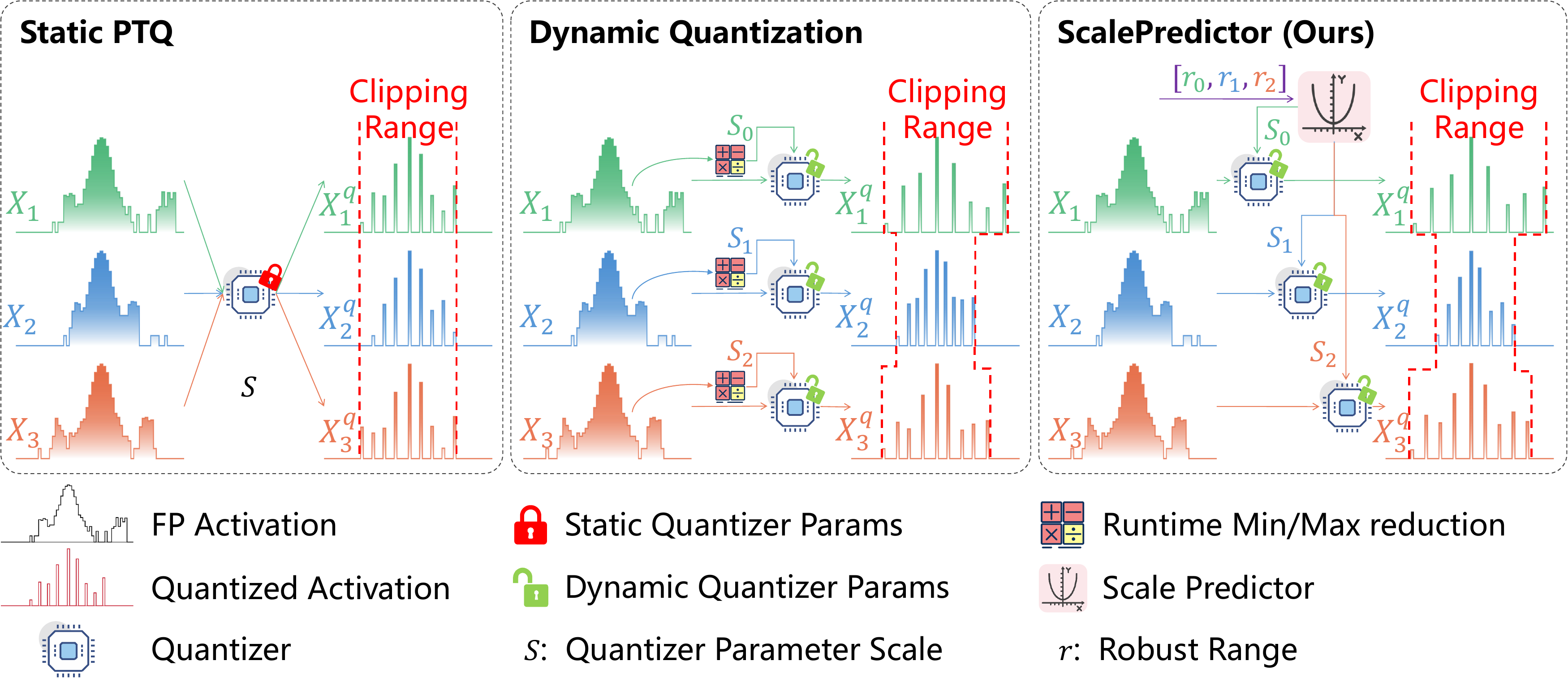}
  \caption{\textbf{Comparison of different quantization paradigms.} (a) Static PTQ: All input samples share a single, unified static quantizer with fixed parameters. (b) Dynamic Quantization: Each sample calculates its own quantizer parameters on-the-fly based on runtime min/max reductions of the current activation tensor. (c) ScalePredictor (Ours): A robust range $r(\mathbf{x})$ is extracted once per image from the patch embedding output, and a scale predictor is used to compute instance-specific quantizer parameters for subsequent layers without per-layer runtime tensor reductions.}
  \label{fig:paradigm_comparison}
\vspace{-1.5em}
\end{figure}

Despite their merits, most existing PTQ methods adopt sample-agnostic quantization configurations, where all input samples share a single set of quantizer parameters, as illustrated in Fig.~\ref{fig:paradigm_comparison}(a). Such designs neglect the substantial diversity among individual instances in natural image datasets, which becomes particularly pronounced under low-bit settings such as W3/A3, where activation heterogeneity across samples can severely degrade performance.
A natural solution is to employ dynamic quantization strategies that adapt quantizer parameters to each input instance. However, existing dynamic quantization techniques~\cite{liu2022instance,moon2024instance} typically calculate quantization parameters on-the-fly via per-layer, full-tensor reductions~\cite{nagel2021white,xiao2023smoothquant}. As illustrated in Fig.~\ref{fig:paradigm_comparison}(b), such runtime reductions introduce severe synchronization barriers and memory-bound overheads on modern hardware accelerators, drastically undermining inference speed.

\begin{figure}[!t]
  \centering
  \includegraphics[width=\textwidth]{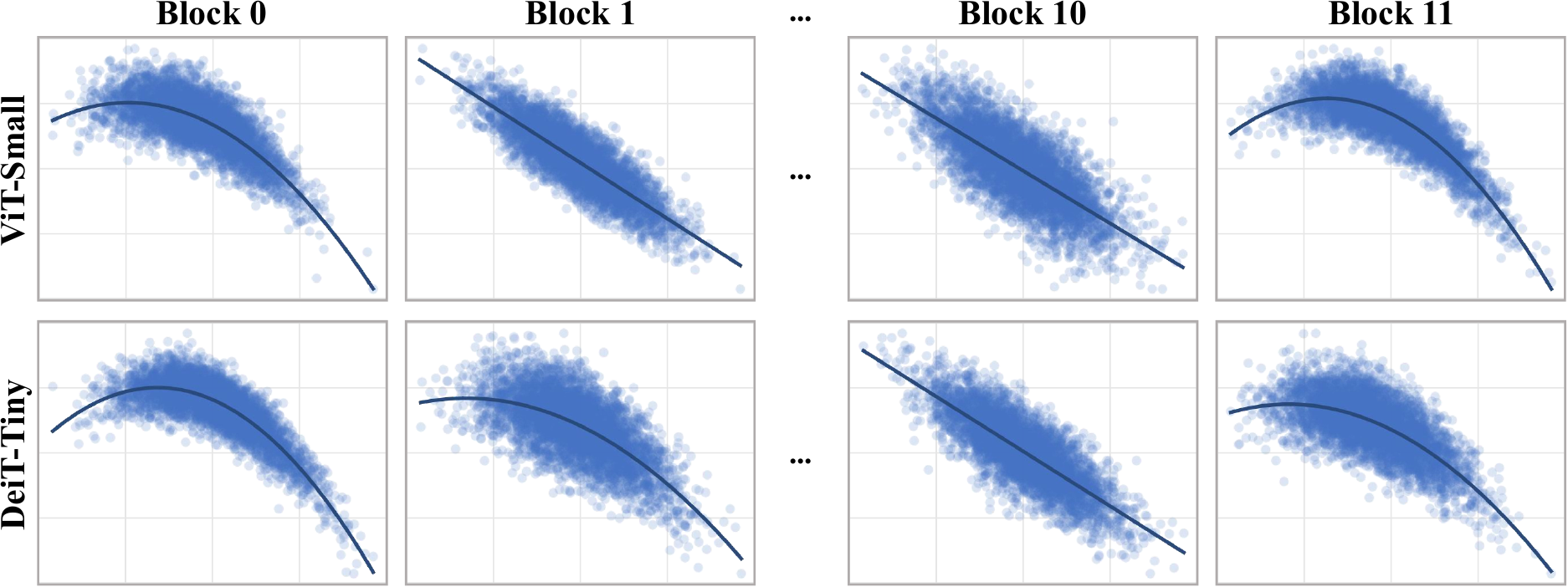}
  \caption{\textbf{Correlation between Shallow Range and Layer-wise Optimal Scales.} It plots the relationship across 5,000 validation samples for different transformer blocks in ViT-Small~\cite{dosovitskiy2020image} and DeiT-Tiny~\cite{touvron2021training}. The x-axis represents the robust range $r(\mathbf{x})$, \ie, the chunk-averaged activation spread of the Patch Embedding output that suppresses outlier influence. The y-axis denotes the oracle per-sample optimal scale $s_l^*(\mathbf{x})$ for the input activation quantization of the corresponding qkv layers at different depths.} \label{fig:scatter}
\vspace{-1em}
\end{figure}

In this work, we revisit instance-level activation heterogeneity from a different perspective. We observe that the optimal activation quantization scales across different transformer layers are not independent. Instead, they exhibit a strong dependency on the activation magnitude observed at shallow layers. As illustrated in Fig.~\ref{fig:scatter}, the robust activation range extracted from the patch embedding output shows a clear and consistent correlation with the oracle per-sample optimal scales across multiple transformer blocks. Notably, this trend persists across both early and late layers, as well as across different model architectures such as ViT-Small and DeiT-Tiny. These observations suggest that instance-level scale variations propagate coherently through the network depth and can be effectively captured by early-layer activation statistics.

Motivated by this insight, we propose a novel post-training quantization method, termed \textbf{ScalePredictor}, for accurate and efficient dynamic compression of ViTs. As illustrated in Fig.~\ref{fig:paradigm_comparison}(c), ScalePredictor predicts instance-specific quantization scales from shallow-layer statistics, avoiding repeated runtime reductions during inference. More specifically, we first introduce a \textbf{robust range extraction} mechanism that computes stable activation statistics from shallow layers, effectively mitigating the influence of outliers without relying on hardware-unfriendly percentile sorting. Building on this statistic, we further design a \textbf{Taylor-Motivated polynomial predictor} module that takes the robust range as input and predicts all layer-wise quantization scales in a single pass.
The projection coefficients are learned during the calibration stage, enabling the model to capture cross-layer scale dependencies with negligible inference overhead. As a result, quantization parameters can be determined early in the inference pipeline, enabling instance-aware quantization while eliminating the need for repeated runtime reductions at every layer.

In conclusion, our main contributions are summarized as follows:
\begin{itemize}
    \item We introduce ScalePredictor, a novel approach designed to address instance-level distribution heterogeneity via instance-aware scale prediction. To the best of our knowledge, ScalePredictor is the first post-training quantization method for ViTs that leverages dynamic quantization.
    \item ScalePredictor integrates a robust range extraction mechanism with a Taylor-inspired polynomial scale projection, enabling accurate quantization scale prediction on-the-fly with minimal inference overhead.
    \item Extensive experiments on ImageNet demonstrate that ScalePredictor achieves consistent improvements over static PTQ baselines across a wide range of quantization regimes, while introducing negligible additional overhead.
\end{itemize}
\vspace{-1em}



\section{Related Work}

\noindent\textbf{Post-Training Quantization.}
Post-training quantization (PTQ) slims neural networks by limiting the bit-width of weights and activations with a few calibration data, which have been recognized as a popular model compression technique. 
Approaches like FQ-ViT~\cite{lin2021fq}, PTQ4ViT~\cite{yuan2022ptq4vit} and RepQ-ViT~\cite{li2023repq} address the post-Softmax/GeLU distribution compatibility, while Bi-ViT~\cite{li2024bivit} and AdaLog~\cite{wu2024adalog} explore binary quantization and logarithmic quantization for extreme compression. To mitigate the quantization errors of model weights, AIQViT~\cite{jiang2025aiqvit} leverages a low-rank compensation mechanism, which determines the ranks by network architecture search. 
Furthermore, SmoothQuant~\cite{xiao2023smoothquant} and Outlier Suppression+~\cite{wei2023outlier} propose migrating quantization difficulty from activations to weights through mathematical equivalence, thereby effectively mitigating the impact of outliers. More recently, GPTQ~\cite{frantar2022gptq} and ZeroQuant~\cite{yao2022zeroquant} extend PTQ techniques to billion-scale language Transformer models, demonstrating the broad applicability of quantization approaches across model scales.
However, most ViT-oriented PTQ methods ignore the diversity of instances, so that they use fixed quantization parameters uniformly. Such a paradigm generally shows limited performance, especially in ultra-low bit configuration.

\noindent\textbf{Dynamic Neural Networks.}
Dynamic neural networks adapt their computation to each input instance at runtime, enabling efficient and instance-aware inference.
In the quantization domain, this dynamism takes several forms. Standard dynamic quantization computes per-layer min/max activation statistics at runtime, achieving instance-specific scales but incurring synchronization overhead that grows linearly with network depth. Instance-aware mixed-precision methods such as DQNet~\cite{liu2022instance} and IGQ-ViT~\cite{moon2024instance} instead assign different bit-widths to layers or samples via a learned policy~\cite{dong2020hawq}; while effective, they require QAT with full training data and specialized mixed-precision hardware.
A closely related work is QPP~\cite{kryzhanovskiy2021qpp}, which learns a regression model to estimate quantization parameters for each input sample prior to inference. Our work shares a similarity with QPP in that both methods attempt to avoid expensive runtime reductions by predicting instance-aware quantization parameters. However, our method differs from QPP in several key aspects: (\textbf{i}) QPP is designed for CNNs and operates on raw image pixels, whereas our method targets Vision Transformers and leverages the Patch Embedding output already computed in the forward pass. (\textbf{ii}) QPP predicts quantization parameters independently for each layer. In contrast, our method predicts all layer-wise scales jointly from a single shallow-layer statistic. (\textbf{iii}) QPP relies on generic regression predictors, while our method introduces a Taylor-inspired polynomial projection to model cross-layer scale dependencies.


\section{Method}
\label{sec:method}

In this section, we present \textbf{ScalePredictor}, an instance-aware post-training quantization framework tailored for Vision Transformers. We first revisit the preliminaries of ViT architecture and the widely-used uniform affine quantization. After that, We present the motivation for our proposed method. Next, we detail the proposed ScalePredictor, involving a robust range extraction approach and a Taylor-motivated polynomial scale projection module. Finally, we give the training process and the efficiency analysis during inference.


\subsection{Preliminaries}
\label{sec:method:prelim}

\noindent\textbf{Vision Transformers.} 
Vision Transformer processes an image by first dividing it into a sequence of patches, which are linearly projected into token embeddings $\mathbf{X}_0 \in \mathbb{R}^{N \times C}$, where $N$ is the number of tokens and $C$ is the embedding dimension. The tokens are then processed by $N_b$ consecutive transformer blocks. Each block consists of a Multi-Head Self-Attention (MSA) module and a Feed-Forward Network (FFN), preceded by Layer Normalization (LN). The forward propagation for the $b$-th block is formulated as:
\begin{align}
    \mathbf{X}'_b &= \mathbf{X}_{b-1} + \mathrm{MSA}(\mathrm{LN}(\mathbf{X}_{b-1})), \label{eq:vit_msa}\\
    \mathbf{X}_b &= \mathbf{X}'_b + \mathrm{FFN}(\mathrm{LN}(\mathbf{X}'_b)). \label{eq:vit_ffn}
\end{align}

\noindent\textbf{Uniform Affine Quantization.}
Uniform affine quantization is widely used to convert the high-precision floating-point  activations (e.g., 32-bit) and weights to low-bit integers (e.g., 8-bit). Given a floating-point activation tensor $\mathbf{X}$, the quantization process is:
\begin{equation}
\label{eq:quantization}
Q(\mathbf{X}; s, z) = s \cdot \mathrm{clip}\left(\left\lfloor \frac{\mathbf{X}}{s} \right\rceil + z, 0, 2^n-1\right) - s \cdot z,
\end{equation}
where $s > 0$ is the quantization scale, $z$ is the zero-point, $n$ is the bit-width, $\lfloor \cdot \rceil$ denotes the round-to-nearest operator, and $\mathrm{clip}(\cdot)$ bounds the values within the integer range. 


\subsection{Motivation}
\label{sec:method:motivation} 
A critical challenge in low-bit quantization of ViTs is the severe variation of activation distributions across different input instances. To address such volatility, standard Dynamic Quantization (DQ) computes the quantization scale on-the-fly for each layer based on the current input tensor, as illustrated in Fig.~\ref{fig:paradigm_comparison}(b):
\begin{equation}
\label{eq:dq_scale}
s_{\mathrm{DQ}}(\mathbf{X}) = \frac{\max(|\mathbf{X}|)}{2^{n-1}-1}.
\end{equation}
Although adapting successfully to instance-specific fluctuations, DQ necessitates a full-tensor reduction over every activation (\eg, $197 \times 768$ elements in ViT-Base) at runtime. Consequently, such operations introduce severe synchronization barriers and substantially increase the overall inference overhead~\cite{luo2025clusterfusion}.

This raises a fundamental question: \textit{How to retain instance-wise adaptivity without incurring massive just-in-time computation.} To answer this, we analyze the oracle optimal scales across various depths offline. As illustrated in Fig.~\ref{fig:scatter}, our investigation reveals a consistent functional dependency where the robust range extracted from shallow activations strongly correlates with the ideal scales of subsequent layers. Such a phenomenon indicates that early network representations natively encapsulate sufficient instance-level context to guide subsequent scaling decisions. Furthermore, it also shows that mapping is not strictly uniform and inherently drifts across different network depths and model architectures, which motivates a {learnable}, layer-aware prediction mechanism that adapts to specific structures during calibration.



\subsection{ScalePredictor}
\label{sec:method:predictor}

\begin{figure}[!t]
  \centering
  \includegraphics[width=\textwidth]{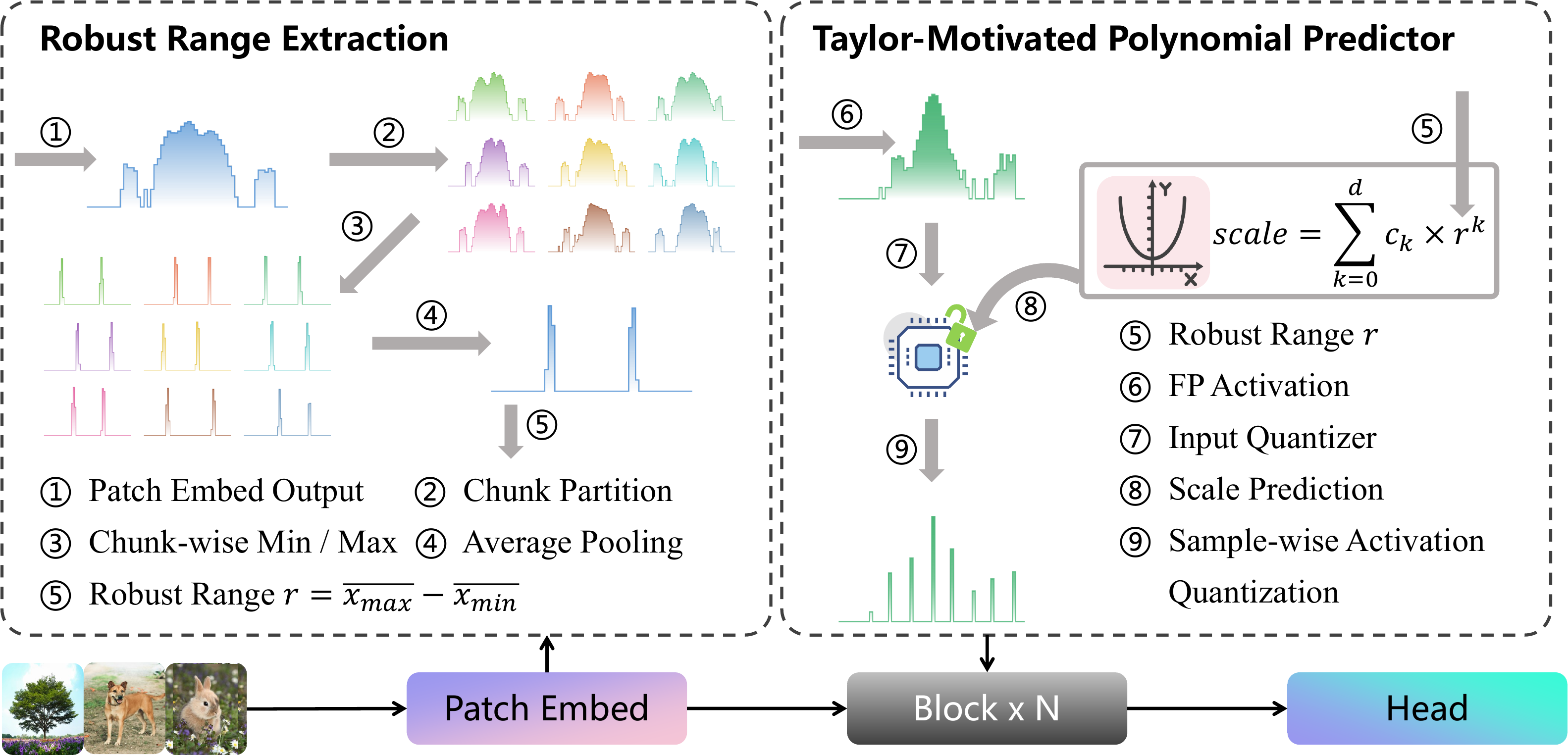}
  \caption{ \textbf{Overview of ScalePredictor.} \textbf{The left panel:} the patch embedding output is partitioned into chunks to compute chunk-wise min/max values, which are then average-pooled to extract a robust range $r(\mathbf{x})$. \textbf{The right panel:} the extracted robust range $r(\mathbf{x})$ is fed into a polynomial scale predictor to generate the sample-wise scale, which configures the input quantizer for the subsequent full-precision activations.}
  \label{fig:pipeline}
  \vspace{-1em}
\end{figure}

In this section, we propose \textbf{ScalePredictor} to seamlessly instantiate this instance-aware scale prediction. The overall pipeline of our method is illustrated in Fig.~\ref{fig:pipeline}. Specifically, it comprises a \textbf{Robust Range Extraction} mechanism to distill a stable scalar context from the initial patch embeddings, and a \textbf{Taylor-Motivated Polynomial Predictor} to model the dependencies between this shallow context and the ideal scales of subsequent layers. 


\noindent\textbf{Robust Range Extraction.}
Inspired by ~\cite{cadena2018diverse} and \cite{kryzhanovskiy2021qpp}, we hypothesize that the deep feature distributions of an individual input image are firmly anchored in its shallow representations. Let $\mathbf{x}$ denote an input image, and let $\mathbf{x}_0 \in \mathbb{R}^{N \times C}$ denote the output of the Patch Embedding layer, \ie, the first feature layer that projects raw image patches into token embeddings before any Transformer block. To avoid the negative influence of outliers, which frequently emerge in ViTs but are largely irrelevant to the global image contrast, we aim to extract a robust range $r(\mathbf{x})$ for each instance. While a straightforward global min/max operation is highly sensitive to isolated activation outliers, and exact quantile-based clipping, for example $99.9\%$, requires time-consuming sorting operations, we propose a balanced alternative termed \textbf{Chunk-based Min/Max Average}. 

Specifically, we first partition the flattened tensor $\mathbf{x}_0$ into $K$ non-overlapped chunks with equal size, denoted as $\{\mathbf{x}_0^{(1)}, \mathbf{x}_0^{(2)}, \dots, \mathbf{x}_0^{(K)}\}$. We then compute the local maximum and minimum for each chunk, and average these local extrema to obtain the robust global maximum and minimum. The robust range $r(\mathbf{x})$ is defined as the difference between these averaged extrema:
\begin{equation}
r(\mathbf{x}) = \frac{1}{K} \sum_{i=1}^{K} \max(\mathbf{x}_0^{(i)}) - \frac{1}{K} \sum_{i=1}^{K} \min(\mathbf{x}_0^{(i)}).
\label{eq:robust_range}
\end{equation}
This chunk-based averaging acts as a coarse-grained percentile approximation. By carefully selecting the number of chunks $K$, it effectively smooths out isolated outliers without excessively degrading the true dynamic range. To ensure numerical stability and prevent out-of-distribution samples from causing the predicted scale to diverge, we clamp $r(\mathbf{x})$ within $[r_{\min}, r_{\max}]$, determined by the $0.1\%$ and $99.9\%$ quantiles of $r(\mathbf{x})$ over the calibration set. For simplicity, we use the clamped context $\hat{r}(\mathbf{x}) = \mathrm{clip}(r(\mathbf{x}),r_{\min},r_{\max})$ and omit the hat in the following.
For a mini-batch, this procedure is applied independently to each image, producing a vector of contexts $\{r(\mathbf{x}_i)\}$, and the subsequent scale prediction yields per-layer, per-image scales $\{s_l(\mathbf{x}_i)\}$. 

The above process is exceptionally efficient, since computing local extrema and their average requires only $\mathcal{O}(N \cdot C)$ complexity and completely avoids sorting. Besides, this extraction is performed exactly once per image at the very beginning of the network.



\noindent\textbf{Taylor-Motivated Polynomial Predictor.}
In a deep ViT,  let $l = 1, \ldots, M$ index all quantized activation operators sequentially across the network. The oracle per-sample activation scale $s_l^*(\mathbf{x})$ for the $l$-th quantized layer is influenced by the input context summarized by $r(\mathbf{x})$. 
However, due to the nature of complicated architectures of ViTs, explicitly modeling the dependency between the optimal scales in deeper layers and the shallow ones is intractable. Instead, we model $s_l^*(\mathbf{x})$ as being governed by a dominant implicit non-linear trend $\Phi_l\big(r(\mathbf{x})\big)$:
\begin{equation}
\label{eq:scale_approx}
s_l^*(\mathbf{x}) \approx \Phi_l\big(r(\mathbf{x})\big)
\end{equation}
After that, another problem arises: \textit{How to effectively learn this trend.} To answer this, we first assume the $\Phi_l\big(r(\mathbf{x})\big)$ is $d$-times differentiable, which can be roughly validated in Fig.~\ref{fig:scatter}. Then, motivated by a local Taylor expansion around the expected context $\bar{r}=\mathbb{E}_{\mathbf{x}\sim\mathcal{D}_c}[r(\mathbf{x})]$ over the calibration set, we reformulate the trend $\Phi_l\big(r(\mathbf{x})\big)$ as:
\begin{equation}
\label{eq:taylor_expand}
s_l(\mathbf{x}) \approx \Phi_l(\bar{r}) + \Phi_l'(\bar{r})\big(r(\mathbf{x}) - \bar{r}\big) + \dots +\frac{1}{d!}\Phi_l^{(d)}(\bar{r})\big(r(\mathbf{x}) - \bar{r}\big)^d + \mathcal{O}\big((r(\mathbf{x}) - \bar{r})^{d+1}\big),
\end{equation}
where $\Phi_l'(\bar{r})$ and $\Phi_l^{(d)}(\bar{r})$ respectively denote the first-order and $d$-order derivative at the point of $\bar{r}$. By expanding the polynomial terms and absorbing the constant $\bar{r}$ along with the derivatives into the learnable parameters, we formulate the scale prediction as a lightweight polynomial:
\begin{equation}
s_l(\mathbf{x}) = \sum_{k=0}^{d} c_{l,k} \cdot (r(\mathbf{x}))^k,
\label{eq:taylor_poly}
\end{equation}
where $\mathbf{c}_l = \{c_{l,0}, \dots, c_{l,d}\}$ are the learnable projection coefficients for the $l$-th quantized layer. This formulation ensures that each individual sample $\mathbf{x}$ receives its own unique quantization scale $s_l(\mathbf{x})$ tailored to its specific shallow activation range $r(\mathbf{x})$. By controlling the polynomial degree $d$, we can effectively balance the model's capacity to capture non-linear scale variations against the risk of overfitting on the limited calibration data. In practice, the optimal degree $d$ can be flexibly selected based on the specific architecture and quantization bit-width to achieve the best trade-off. Notably, we find that even low-degree polynomials are sufficient to achieve strong performance, as demonstrated in Sec.~\ref{sec:experiments}.





\subsection{Calibration and Hardware-Friendly Inference}
\label{sec:method:optimization}

\noindent\textbf{Calibration Objective.}
In the calibration phase, we first obtain the base quantization parameters of the activation quantizers using standard sample-agnostic calibration methods (\eg, Min-Max, grid search, or learning-based optimization over the calibration set). To stabilize the training process, we initialize the constant term $c_{l,0}$ of the predictor with the obtained optimal sample-agnostic scale, and set the higher-order coefficients (\eg, $c_{l,1}, c_{l,2}$) to zero. This guarantees that the dynamic predictor initially behaves exactly like a well-calibrated sample-agnostic quantizer. We then optimize the polynomial coefficients block by block. For the $b$-th transformer block, we minimize the block-wise reconstruction loss~\cite{li2021brecq} on the calibration set $\mathcal{D}_c$:
\begin{equation}\label{eq:recon_loss}
\min_{\{\mathbf{c}_l\}_{l \in \mathcal{I}_b}} \mathbb{E}_{\mathbf{X} \sim \mathcal{D}_c} \left\| \mathcal{F}_b(\mathbf{X}_{b-1}) - \mathcal{F}_b^q(\mathbf{X}_{b-1}^q; \{\mathbf{c}_l\}_{l \in \mathcal{I}_b}, \mathbf{r}(\mathbf{X})) \right\|_F^2,
\end{equation}

\noindent\textbf{Complexity and Inference Overhead.}
During inference, ScalePredictor computes the robust range $r(\mathbf{x})$ once per image at the Patch Embedding layer via an $\mathcal{O}(N \cdot C)$ chunk-based reduction. For every subsequent quantized layer, the instance-specific scale $s_l(\mathbf{x})$ is produced by evaluating a degree-$d$ polynomial (Eq.~\ref{eq:taylor_poly}), which costs $\mathcal{O}(1)$ FLOPs and requires no access to deep activation tensors. By contrast, conventional Dynamic Quantization generally performs an $\mathcal{O}(N \cdot C)$ min/max reduction per layer, per instance, creating a significant memory-bound bottleneck. Our method therefore achieves the same per-layer inference complexity as static PTQ, whose activation scales are fixed at calibration time, while preserving instance-level adaptivity. We empirically validate in Sec.~\ref{sec:efficiency} that this translates to only 0.4--0.8\% end-to-end latency overhead over static PTQ on an RTX~3090 GPU. The complete calibration and inference pipeline is summarized in Algorithm~\ref{alg:scalepredictor}.

\begin{algorithm}[!htb]
\caption{ScalePredictor: Calibration and Inference}
\label{alg:scalepredictor}
\textbf{Input:} Full-precision ViT $\mathcal{M}$, Calibration set $\mathcal{D}_c$, Bit-width $n$, Polynomial degree $d$ \\
\textbf{Output:} Quantized ViT $\mathcal{M}_q$ with optimized polynomial coefficients $\mathbf{c}$

\begin{algorithmic}[1]
\STATE \textbf{\% Phase 1: Offline Calibration}
\FOR{each transformer block $b = 1, 2, \dots, N_b$}
  \FOR{each quantized layer $l \in \mathcal{I}_b$}
    \STATE Initialize $c_{l,0}$ with the sample-agnostic optimal scale for layer $l$, and set $c_{l,k} \leftarrow 0$ for $k > 0$.
  \ENDFOR
    \WHILE{not converged}
        \STATE Sample a mini-batch of images $\mathbf{X} \sim \mathcal{D}_c$.
        \STATE Get Patch Embedding output $\mathbf{X}_0$ and compute a vector of robust ranges $\mathbf{r}$ (one per instance) via Eq.~\ref{eq:robust_range}.
    \STATE Clamp $\mathbf{r}$ element-wise to $[r_{\min}, r_{\max}]$.
        \FOR{each quantized layer $l \in \mathcal{I}_b$}
          \STATE Predict instance-wise scales $\mathbf{s}_l = \sum_{k=0}^{d} c_{l,k} \cdot \mathbf{r}^k$.
        \ENDFOR
        \STATE Compute block-wise reconstruction loss $\mathcal{L}$ via Eq.~\ref{eq:recon_loss}.
        \STATE Update $\{\mathbf{c}_l\}_{l \in \mathcal{I}_b}$ using gradient descent with the Straight-Through Estimator.
    \ENDWHILE
\ENDFOR
\STATE \textbf{\% Phase 2: Online Inference}
\STATE \textbf{Given:} A single new input image $\mathbf{x}$
\STATE Compute Patch Embedding output $\mathbf{x}_0$ from $\mathbf{x}$.
\STATE Extract instance-specific scale context $r(\mathbf{x})$ from $\mathbf{x}_0$ using Eq.~\ref{eq:robust_range}.
\STATE Clamp $r(\mathbf{x})$ to $[r_{\min}, r_{\max}]$.
\FOR{each transformer block $b = 1, 2, \dots, N_b$}
  \FOR{each quantized layer $l \in \mathcal{I}_b$}
    \STATE Predict instance scale $s_l(\mathbf{x}) = \sum_{k=0}^{d} c_{l,k} \cdot r(\mathbf{x})^k$ \hfill \textit{\% $\mathcal{O}(1)$ scalar op}
  \ENDFOR
  \STATE Execute quantized block forward pass $\mathcal{F}_b^q$ using $\{s_l(\mathbf{x})\}_{l \in \mathcal{I}_b}$.
\ENDFOR
\end{algorithmic}
\end{algorithm}

\section{Experiments}
\label{sec:experiments}

\subsection{Experimental Setup}
\textbf{Datasets and Models.} We evaluate our proposed ScalePredictor on the ImageNet dataset~\cite{deng2009imagenet}, which contains 1.2 million training images and 50,000 validation images across 1,000 classes. We conduct extensive experiments on a wide range of representative Vision Transformer architectures, including ViT (ViT-Small, ViT-Base)~\cite{dosovitskiy2020image}, DeiT (DeiT-Tiny, DeiT-Small, DeiT-Base)~\cite{touvron2021training}, and Swin Transformer (Swin-Small, Swin-Base)~\cite{liu2021swin}.

\noindent \textbf{Baselines.} To demonstrate the effectiveness of our method, we integrate ScalePredictor into several representative and widely adopted post-training quantization frameworks, including classical PTQ methods like BRECQ~\cite{li2021brecq} and QDrop~\cite{wei2022qdrop}, and  ViT-tailored approaches like AdaLog~\cite{wu2024adalog} and I\&S-ViT~\cite{zhong2023s}. 

\begin{table*}[!t]
\centering
\footnotesize
\setlength{\tabcolsep}{3pt}
\renewcommand{\arraystretch}{1.12}
\caption{\textbf{Experimental results on ImageNet.} Top-1 accuracy (\%) across seven ViT architectures (ViT-S/B, DeiT-T/S/B, Swin-S/B) under three bit-width configurations (W2/A3, W3/A3, W4/A3). We compare ScalePredictor (Ours) integrated into four PTQ baselines (BRECQ, QDrop, AdaLog, I\&S-ViT) against: (i)~the baselines themselves with static scales, and (ii)~dynamic quantization methods (JIT Min/Max and JIT Percentile), which compute activation scales on-the-fly via runtime reductions. 
}
\label{tab:main_comparison}
\resizebox{\textwidth}{!}{%
\begin{tabular}{c l r r r r r r r}
\toprule
 Bits (W/A) &  Method & ViT-S & ViT-B & DeiT-T & DeiT-S & DeiT-B & Swin-S & Swin-B \\
\midrule
 \multirow{10}{*}{W2/A3}                     &JIT Min/Max~\cite{nagel2021white}  &  0.17 &  0.14 &  0.30 &  0.20 &  0.30 &  0.19 &  0.19 \\
& JIT Percentile~\cite{nagel2021white}                          &  0.50 &  0.62 &  1.65 &  0.97 & 18.16 &  3.40 &  3.95 \\
&  BRECQ~\cite{li2021brecq}                               &  1.10 & 17.58 &  2.23 &  5.00 & 43.34 &  3.14 & 13.16 \\
& \cellcolor{gray!25}BRECQ~+~Ours &  \cellcolor{gray!25} 4.15 & \cellcolor{gray!25}24.56 & \cellcolor{gray!25} 4.99 & \cellcolor{gray!25}10.36 & \cellcolor{gray!25}48.96 & \cellcolor{gray!25} 4.33 & \cellcolor{gray!25}20.36 \\
& QDrop~\cite{wei2022qdrop}                               &  7.75 & 30.61 &  9.87 &  3.99 & 48.72 & 21.97 & 40.23 \\
& \cellcolor{gray!25}QDrop~+~Ours &  \cellcolor{gray!25}10.49 & \cellcolor{gray!25}34.98 & \cellcolor{gray!25}12.29 & \cellcolor{gray!25} 9.43 & \cellcolor{gray!25}51.36 & \cellcolor{gray!25}20.19 & \cellcolor{gray!25}42.42 \\
  &  AdaLog~\cite{wu2024adalog}                           &  2.34 &  9.37 &  5.70 & 12.08 & 39.99 &  6.44 &  2.93 \\
& \cellcolor{gray!25}AdaLog~+~Ours &  \cellcolor{gray!25} 3.02 & \cellcolor{gray!25}11.00 & \cellcolor{gray!25} 7.25 & \cellcolor{gray!25}12.58 & \cellcolor{gray!25}42.40 & \cellcolor{gray!25} 8.95 & \cellcolor{gray!25}13.05 \\
 &   I\&S-ViT~\cite{zhong2023s}                           & 16.10 & 53.09 & 14.51 & 23.94 & 61.61 & 33.57 & 45.82 \\
& \cellcolor{gray!25}I\&S-ViT~+~Ours &  \cellcolor{gray!25}16.79 & \cellcolor{gray!25}53.51 & \cellcolor{gray!25}19.75 & \cellcolor{gray!25}30.31 & \cellcolor{gray!25}63.79 & \cellcolor{gray!25}44.15 & \cellcolor{gray!25}51.87 \\
\cmidrule(lr){1-9}
\multirow{10}{*}{W3/A3} &JIT Min/Max~\cite{nagel2021white}                     &  0.31 &  0.16 &  0.32 &  0.38 &  0.35 &  0.25 &  0.20 \\
& JIT Percentile~\cite{nagel2021white}                           &  4.79 &  4.31 &  8.88 &  4.54 & 42.51 & 40.19 & 41.64 \\
 & AIQViT~\cite{jiang2025aiqvit} &41.32 &43.68 &38.51 &55.36 &66.15  &71.42 &63.01 \\
& BRECQ~\cite{li2021brecq}                                & 31.46 & 53.92 & 22.76 & 34.75 & 68.20 & 56.03 & 61.49 \\
& \cellcolor{gray!25}BRECQ~+~Ours  & \cellcolor{gray!25}40.17 & \cellcolor{gray!25}58.41 & \cellcolor{gray!25}30.62 & \cellcolor{gray!25}41.98 & \cellcolor{gray!25}70.26 & \cellcolor{gray!25}57.16 & \cellcolor{gray!25}64.09 \\
& QDrop~\cite{wei2022qdrop}                               & 28.93 & 59.90 & 38.58 & 32.60 & 69.22 & 62.29 & 66.42 \\
& \cellcolor{gray!25}QDrop~+~Ours &  \cellcolor{gray!25}34.00 & \cellcolor{gray!25}62.63 & \cellcolor{gray!25}39.12 & \cellcolor{gray!25}43.00 & \cellcolor{gray!25}69.94 & \cellcolor{gray!25}61.36 & \cellcolor{gray!25}67.07 \\
&  AdaLog~\cite{wu2024adalog}                             & 37.46 & 60.16 & 34.45 & 49.28 & 67.55 & 63.01 & 63.84 \\
& \cellcolor{gray!25}AdaLog~+~Ours &  \cellcolor{gray!25}42.61 & \cellcolor{gray!25}60.97 & \cellcolor{gray!25}37.37 & \cellcolor{gray!25}51.35 & \cellcolor{gray!25}68.80 & \cellcolor{gray!25}64.78 & \cellcolor{gray!25}66.32 \\
& I\&S-ViT~\cite{zhong2023s}                              & 47.38 & 71.20 & 44.50 & 58.89 & 73.79 & 68.66 & 70.86 \\
& \cellcolor{gray!25}I\&S-ViT~+~Ours &  \cellcolor{gray!25}48.50 & \cellcolor{gray!25}71.53 & \cellcolor{gray!25}48.10 & \cellcolor{gray!25}62.03 & \cellcolor{gray!25}74.59 & \cellcolor{gray!25}70.95 & \cellcolor{gray!25}72.85 \\
\cmidrule(lr){1-9}
 \multirow{10}{*}{W4/A3} & JIT Min/Max~\cite{nagel2021white}        &  0.30 &  0.17 &  0.35 &  0.45 &  0.40 &  0.23 &  0.22 \\
& JIT Percentile~\cite{nagel2021white}                           &  7.97 &  9.28 & 13.41 &  7.50 & 50.24 & 46.14 & 46.55 \\
  & BRECQ~\cite{li2021brecq}                              & 45.95 & 64.70 & 32.32 & 54.72 & 73.54 & 66.04 & 70.96 \\
 & \cellcolor{gray!25}BRECQ~+~Ours & \cellcolor{gray!25}51.71 & \cellcolor{gray!25}67.04 & \cellcolor{gray!25}39.79 & \cellcolor{gray!25}58.54 & \cellcolor{gray!25}73.90 & \cellcolor{gray!25}66.55 & \cellcolor{gray!25}72.02 \\
 & QDrop~\cite{wei2022qdrop}                              & 37.19 & 67.88 & 46.46 & 58.07 & 73.91 & 69.11 & 72.54 \\
&  \cellcolor{gray!25}QDrop~+~Ours & \cellcolor{gray!25}41.33 & \cellcolor{gray!25}68.90 & \cellcolor{gray!25}45.73 & \cellcolor{gray!25}60.74 & \cellcolor{gray!25}74.38 & \cellcolor{gray!25}68.55 & \cellcolor{gray!25}72.46 \\
 & AdaLog~\cite{wu2024adalog}                             & 51.03 & 69.12 & 44.74 & 63.60 & 74.12 & 71.57 & 73.23 \\
& \cellcolor{gray!25}AdaLog~+~Ours  & \cellcolor{gray!25}55.06 & \cellcolor{gray!25}69.11 & \cellcolor{gray!25}47.55 & \cellcolor{gray!25}65.84 & \cellcolor{gray!25}74.56 & \cellcolor{gray!25}72.11 & \cellcolor{gray!25}74.28 \\
& I\&S-ViT~\cite{zhong2023s}                              & 57.45 & 75.67 & 52.85 & 68.71 & 77.17 & 73.51 & 75.25 \\
& \cellcolor{gray!25}I\&S-ViT~+~Ours &  \cellcolor{gray!25}58.14 & \cellcolor{gray!25}75.77 & \cellcolor{gray!25}55.43 & \cellcolor{gray!25}69.67 & \cellcolor{gray!25}77.27 & \cellcolor{gray!25}74.80 & \cellcolor{gray!25}76.37 \\
\bottomrule
\end{tabular}%
}
\vspace{-1em}
\end{table*}

\noindent \textbf{Implementation Details.} Our implementation is based on the publicly available codebase of I\&S-ViT~\cite{zhong2023s}. Following standard PTQ protocols, we randomly sample 1,024 images from the ImageNet training set for calibration. We use symmetric uniform quantization for weights. Activation quantization follows each baseline (Log-$\sqrt{2}$ for I\&S-ViT, AdaLog quantization for AdaLog, and uniform quantization otherwise). We extract the robust range $r(\mathbf{x})$ from the Patch Embedding layer using Chunk-based Min/Max Average with $K=32$ chunks. The polynomial predictor is optimized with the straight-through estimator during block-wise reconstruction. ScalePredictor targets activation quantization; weight quantization follows the underlying PTQ baseline. All experiments are conducted on an NVIDIA GeForce RTX 3090 GPU using PyTorch.


\subsection{Main Results}

Table~\ref{tab:main_comparison} summarizes the ImageNet Top-1 accuracy under W2/A3, W3/A3, and W4/A3 configurations. For each PTQ baseline, the unshaded row reports the static quantization result with a degree-0 predictor, \ie a per-layer constant scale, while the shaded ``+~Ours'' row reports the result using our ScalePredictor.

\noindent \textbf{Comparison with PTQ Baselines.}
Integrating ScalePredictor improves accuracy in the vast majority of settings across all four PTQ baselines and seven ViT architectures.
The gains are most pronounced at W2/A3, where activation scale mismatch is most severe: for instance, I\&S-ViT on Swin-Small improves from 33.57\% to 44.15\%, a gain of 10.58\%.
Substantial gains persist at higher precisions, such as BRECQ on ViT-Small from 31.46\% to 40.17\% at W3/A3 and QDrop on DeiT-Small from 32.60\% to 43.00\% at W3/A3. As weight precision increases to W4, the margins naturally shrink since the static baseline already better approximates the optimal scale, yet remain consistently positive for the majority of configurations.
We observe a small number of regressions concentrated in QDrop, the largest being Swin-Small at W2/A3 declining from 21.97\% to 20.19\%; at W4/A3, QDrop also drops slightly on DeiT-Tiny, Swin-Small, and Swin-Base. All regressions remain within 1.8\%, suggesting that the interaction between ScalePredictor and QDrop's dropout-based calibration can occasionally be counterproductive. In extremely challenging cases where the baseline nearly collapses, such as BRECQ on ViT-Small at W2/A3 with only 1.10\%, ScalePredictor still raises accuracy to 4.15\%, demonstrating meaningful relative improvement even when absolute performance remains low.

\noindent \textbf{Comparison with JIT Dynamic Quantization.}
We additionally compare against just-in-time dynamic quantization, abbreviated as JIT, which computes activation scales on-the-fly at inference. From Table~\ref{tab:main_comparison}, we can easily observe that both JIT Min/Max and JIT Percentile severely degrade accuracy compared to PTQ baselines, even though they adapt scales per instance. This is because overriding scales at test time breaks the calibrated weight--activation relationship established during PTQ. In contrast, our ScalePredictor learns instance-aware scales during calibration via a differentiable polynomial predictor with a robust range extraction process, preserving calibration consistency while enabling instance-level adaptivity.

\subsection{Ablation Studies}
\label{sec:ablation}
We conduct ablation studies to examine the key design choices of ScalePredictor, including the predictor input, range extraction location, chunk size and polynomial degree, and predictor function family. All experiments use the W3/A3 quantization setting. 

\begin{wraptable}{r}{0.4\textwidth}
\centering
\vspace{-3em}
\caption{\textbf{Ablation on predictor input} (ViT-S, AdaLog, W3/A3). ImageNet Top-1 accuracy~(\%). \label{tab:ablation_input_feature}
}
\resizebox{\linewidth}{!}{
\begin{tabular}{l c | l c}
\toprule
Input & Acc. & Input & Acc. \\
\midrule
  \cellcolor{gray!25}$r$ & \cellcolor{gray!25}\textbf{42.61} & $\sigma$   & 39.08 \\
  $\mu$      & 37.12          & $\kappa$   & 39.35 \\
\midrule
  $r$+$\sigma$  & 42.38 & $\mu$+$\kappa$     & 40.00 \\
  $r$+$\mu$     & 41.25 & $\sigma$+$\kappa$  & 39.56 \\
  $r$+$\kappa$  & 41.03 & $\mu$+$\sigma$     & 38.34 \\
\midrule
  $r$+$\mu$+$\sigma$      & 41.54 & $\mu$+$\sigma$+$\kappa$      & 39.64 \\
  $r$+$\sigma$+$\kappa$   & 41.34 & $r$+$\mu$+$\sigma$+$\kappa$ & 40.97 \\
  $r$+$\mu$+$\kappa$      & 41.21 & -                          & - \\
\bottomrule
\end{tabular}}
\vspace{-2em}
\end{wraptable}
\noindent \textbf{Influence of Predictor Input.}
We study how different Patch Embedding statistics used as predictor inputs affect quantization performance, including the robust range ($r$), channel-wise mean~($\mu$), standard deviation~($\sigma$), and kurtosis~($\kappa$), and their combinations.
Table~\ref{tab:ablation_input_feature} shows that using the robust range $r$ alone achieves the best accuracy of 42.61\%. Adding extra statistics does not improve performance: the best two-feature combination ($r + \sigma$) reaches 42.38\%, suggesting mild overfitting under limited calibration data.
Combinations that exclude range perform worse, confirming that the robust range captures the most essential input-dependent information for scale prediction.
We therefore adopt range as the sole predictor input for its simplicity and optimality.

\begin{wraptable}{r}{0.4\textwidth}
\centering
\vspace{-3em}
\caption{\textbf{Ablation on predictor function} (DeiT-Tiny, W3/A3). ImageNet Top-1 accuracy (\%). 
}
\label{tab:ablation_mlp}
\small
\begin{tabular}{l c c}
\toprule
Predictor & AdaLog & I\&S-ViT \\
\midrule
MLP               & 33.53 & 44.12 \\
\rowcolor{gray!25} Polynomial & \textbf{37.37} & \textbf{48.10} \\
\bottomrule
\end{tabular}
\vspace{-2em}
\end{wraptable}
\noindent \textbf{Influence of Predictor Function.}
Table~\ref{tab:ablation_mlp} compares our polynomial predictor against a single-hidden-layer MLP (192 hidden units).
The polynomial outperforms the MLP by 3.8-4.0\% while requiring only $d{+}1$ coefficients per quantizer, far fewer than the MLP's hundreds of weights. The polynomial also benefits from a stronger inductive bias aligned with the smooth range-to-scale relationship, avoids overfitting on the limited calibration set, and requires only $\mathcal{O}(d)$ multiply-adds evaluable via Horner's method, confirming that it is both more accurate and more efficient than an MLP alternative.

\begin{figure}[!tb]
\centering
\begin{subfigure}[t]{0.49\linewidth}
\centering
\includegraphics[width=\linewidth]{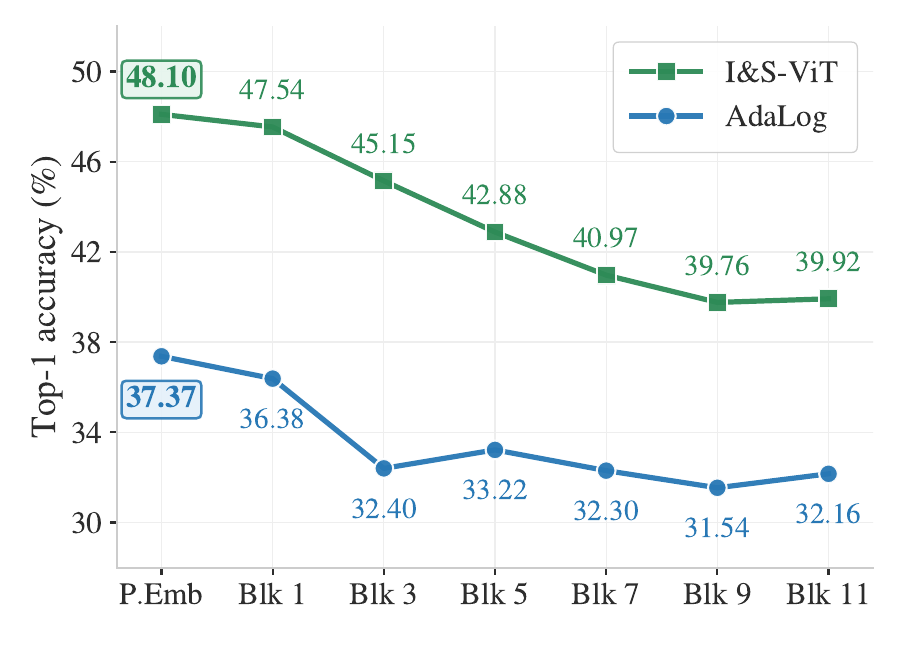}
\caption{Range extraction location.}
\label{fig:ablation_range_location}
\end{subfigure}%
\hfill%
\begin{subfigure}[t]{0.49\linewidth}
\centering
\includegraphics[width=\linewidth]{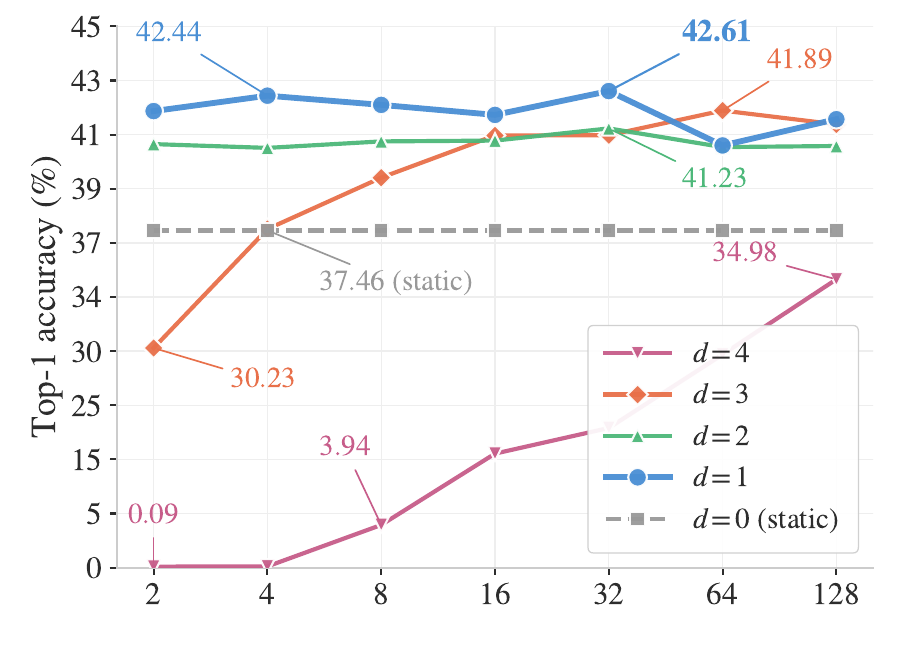}
\caption{Chunk size $K$ and polynomial degree $d$.}
\label{fig:ablation_chunk_degree}
\end{subfigure}
\caption{\textbf{Ablation studies on ViT-S (W3/A3).} (a)~Top-1 accuracy when extracting the robust range from different layers, on AdaLog and I\&S-ViT. (b)~Top-1 accuracy vs.\ chunk size $K$ for polynomial degrees $d{=}0$--$4$ on AdaLog. $d{=}0$ is the static baseline.}
\label{fig:ablation_charts}
\end{figure}

\noindent \textbf{Influence of Range Extraction Location.}
Fig.~\ref{fig:ablation_range_location} compares extracting the robust range from Patch Embedding and several intermediate Transformer blocks (Block~1-11). Patch Embedding consistently performs best. For I\&S-ViT, extracting the range at Patch Embedding achieves 48.10\%, while using Block~11 drops to 39.92\%. A similar trend is observed for AdaLog. Overall, accuracy generally decreases when the extraction point moves deeper, supporting our choice of using shallow features for stable instance-aware context.
We therefore adopt Patch Embedding as the default extraction point. 
\noindent \textbf{Influence of Chunk Size and Polynomial Degree.}
Fig.~\ref{fig:ablation_chunk_degree} jointly studies the chunk size $K$ and polynomial degree $d$. The degree-0 baseline achieves 37.46\%. Degree-1 and degree-2 predictors consistently surpass it by 3--5\% across all $K$ values, peaking at 42.61\% ($K{=}32,\,d{=}1$). Higher degrees become increasingly sensitive to $K$: $d{=}3$ degrades sharply at small $K$ but stabilizes once $K{\geq}8$, while $d{=}4$ never exceeds the static baseline, indicating overfitting under limited calibration data. We therefore adopt $K{=}32$ with $d{\leq}2$ as the default.

\subsection{Efficiency Analysis}
\label{sec:efficiency}

To empirically validate the hardware efficiency of ScalePredictor, we implement the critical inference operations as optimized C++/CUDA kernels compiled via the PyTorch C++ extension. We profile a representative Swin-Small forward pass on an NVIDIA RTX 3090 GPU using CUDA events (30 warmup $+$ 100 timed iterations). We compare four inference modes: (i)~Static PTQ, (ii)~JIT Min/Max (per-layer runtime min/max), (iii)~JIT Percentile (per-layer runtime sort-based 0.1\%/99.9\% quantile), and (iv)~ScalePredictor (Ours): chunk-based range extraction $+$ polynomial scale prediction, once at the Patch Embedding layer). All quantized modes use half-precision (FP16) Tensor Core arithmetic for linear layers, following the standard reduced-precision deployment pipeline.

\begin{table}[!tbp]
\centering
\caption{\textbf{End-to-end inference latency} (ms) of Swin-Small on an RTX 3090 GPU (averaged over 100 runs). 
All modes use FP16 Tensor Core arithmetic for linear layers. $\Delta$ reports the average latency overhead relative to Static PTQ across all batch sizes.}
\label{tab:e2e_latency}
\setlength{\tabcolsep}{4pt}
\small
\begin{tabular}{l r r r c}
\toprule
Method & BS\,=\,16 & BS\,=\,32 & BS\,=\,64 & $\Delta$ \\
\midrule
Static PTQ            & 23.3 & 44.6 &  87.2 & - \\
JIT Min/Max            & 27.8 & 53.3 & 104.0 & +19\% \\
JIT Percentile        & 58.6 & 113.6 & 223.5 & +154\% \\
\rowcolor{gray!25} ScalePredictor (Ours) & 23.5 & 44.8 & 87.6 & +0.6\% \\
\bottomrule
\end{tabular}
\vspace{-0.5em}
\end{table}

\noindent\textbf{End-to-End Latency.}
Table~\ref{tab:e2e_latency} compares end-to-end forward-pass latency across four inference modes at batch sizes 16, 32, and 64. As shown in the $\Delta$ column, JIT Min/Max incurs an average overhead of +19\% due to per-layer min/max reductions, while JIT Percentile is +154\% slower, as per-layer sort operations scale poorly with both layer count and batch size. ScalePredictor introduces only +0.6\% overhead relative to Static PTQ across all batch sizes, confirming that the one-time chunk-based range extraction and polynomial evaluation introduce negligible overhead. This directly validates our core claim: ScalePredictor matches the efficiency of static while delivering the accuracy of dynamic quantization.

\noindent\textbf{Scale Computation Overhead.}
Table~\ref{tab:scale_overhead} isolates the scale computation cost, the only component that differs between static, JIT, and our method. Static PTQ computes scales offline, so its runtime cost is zero. JIT Min/Max requires a full-tensor min/max reduction at every quantized layer; JIT Percentile additionally performs a sort. Our method performs a single chunk-based range extraction at the Patch Embedding layer and then evaluates a degree-2 polynomial for all quantizer points.
Our scale computation is \textbf{40--52$\times$ faster} than JIT Min/Max and \textbf{507--1025$\times$ faster} than JIT Percentile. The advantage grows with batch size because per-layer reductions of JIT methods scale linearly with the batch size, whereas our $\mathcal{O}(1)$ polynomial evaluation is essentially constant.

\begin{table}[!tb]
\begin{minipage}[t]{0.48\linewidth}
\centering
\caption{\textbf{Scale computation overhead} ($\mu$s) of Swin-Small on RTX 3090 GPU (averaged over 200 runs), isolating the cost of computing quantization scales. }
\label{tab:scale_overhead}
\vspace{-1em}
{\scriptsize\setlength{\tabcolsep}{3pt}
\begin{tabular}{l r r r}
\toprule
Method & BS\,=\,16 & BS\,=\,32 & BS\,=\,64 \\
\midrule
JIT Min/Max & 5{,}206 & 7{,}595 & 12{,}250 \\
JIT Percentile & 66{,}136 & 125{,}474 & 243{,}882 \\
\rowcolor{gray!25} ScalePredictor (Ours) & 130 & 150 & 238 \\
\bottomrule
\end{tabular}}
\end{minipage}%
\hspace{0.04\linewidth}%
\begin{minipage}[t]{0.48\linewidth}
\centering
\caption{\textbf{Range Extraction Latency} ($\mu$s) of different range extraction methods on Swin-Small (RTX 3090 GPU, averaged over 200 runs).}
\label{tab:range_extraction}
\vspace{-1em}
{\scriptsize\setlength{\tabcolsep}{3pt}
\begin{tabular}{l r r r}
\toprule
Method & BS\,=\,16 & BS\,=\,32 & BS\,=\,64 \\
\midrule
Min/Max             & 82 & 136 & 242 \\
Percentile  & 1{,}148 & 2{,}207 & 4{,}352 \\
\rowcolor{gray!25} Chunk & 89 & 134 & 219 \\
\bottomrule
\end{tabular}}
\end{minipage}
\vspace{-1.5em}
\end{table}

\noindent\textbf{Chunk-based vs.\ Percentile Range Extraction.}
Table~\ref{tab:range_extraction} disentangles the range extraction step at the Patch Embedding layer. Plain Min/Max is fastest with $\mathcal{O}(N \cdot C)$ single-pass complexity, but as shown in Sec.~\ref{sec:ablation}, it is susceptible to outliers and yields lower accuracy. Our chunk-based method is $\mathcal{O}(N \cdot C)$ by averaging local extrema over $K{=}32$ chunks, achieving comparable latency to Min/Max while significantly improving robustness. Percentile sorting has $\mathcal{O}(N \cdot C \log(N \cdot C))$ complexity and is \textbf{13--20$\times$} slower than our chunk-based method, making it impractical for deployment.
\vspace{-0.5em}

\section{Conclusion}
In this paper, we propose ScalePredictor, a novel dynamic quantization framework that significantly improves post-training quantization of Vision Transformers in low-bit regimes. By revisiting the hidden relationships between the shallow layer distribution and the optimal scales of subsequent activations, ScalePredictor employs a lightweight polynomial predictor to generate quantization scales independently for each input, leveraging shallow features without introducing heavy computational overhead. Based on this, ScalePredictor can be seamlessly integrated into mainstream PTQ frameworks to boost their performance. Extensive experiments on ImageNet demonstrate that ScalePredictor consistently improves four representative PTQ baselines across seven ViT architectures, with gains up to +10.58\% on I\&S-ViT (Swin-Small, W2/A3) and +8.71\% on BRECQ (ViT-Small, W3/A3), while introducing negligible scale-computation overhead compared to static quantization.  



%
%
\bibliographystyle{splncs04}
\bibliography{main}
\end{document}